Long Title: **Segment Anything for comprehensive analysis of grapevine cluster architecture and berry properties**
Short Title: **Grapevine cluster analysis with SAM**

Efrain Torres-Lomas[1], Jimena Lado-Jimena[2], Guillermo Garcia-Zamora[1], Luis Diaz-Garcia[1]*
[1] Department of Viticulture and Enology, University of California Davis, Davis, CA, 95616, USA
[2] Soil and Water Department, Universidad de la Republica, Montevideo, 11400, Uruguay
*corresponding author: diazgarcia@ucdavis.edu

## Abstract

Grape cluster architecture and compactness are complex traits influencing disease susceptibility, fruit quality, and yield. Evaluation methods for these traits include visual scoring, manual methodologies, and computer vision, with the latter being the most scalable approach. Most of the existing computer vision approaches for processing cluster images often rely on conventional segmentation or machine learning with extensive training and limited generalization. The Segment Anything Model (SAM), a novel foundation model trained on a massive image dataset, enables automated object segmentation without additional training. This study demonstrates out-of-the-box SAM's high accuracy in identifying individual berries in 2D cluster images. Using this model, we managed to segment approximately 3,500 cluster images, generating over 150,000 berry masks, each linked with spatial coordinates within their clusters. The correlation between human-identified berries and SAM predictions was very strong (Pearson's $r^2$=0.96). Although the visible berry count in images typically underestimates the actual cluster berry count due to visibility issues, we demonstrated that this discrepancy could be adjusted using a linear regression model (adjusted $R^2$=0.87). We emphasized the critical importance of the angle at which the cluster is imaged, noting its substantial effect on berry counts and architecture. We proposed different approaches in which berry location information facilitated the calculation of complex features related to cluster architecture and compactness. Finally, we discussed SAM's potential integration into currently available pipelines for image generation and processing in vineyard conditions.

## Introduction

Grape cluster architecture and compactness are important fruit traits that influence yield, quality, and susceptibility to pests and diseases [1]. Cluster architecture is directly related to cluster compactness, which describes the ratio between the volume occupied by berries and the total cluster volume [2]. In other words, cluster architecture determines the arrangement of berries in a cluster and the distribution of free space. Cluster architecture is complex, difficult to measure quantitatively, and determined by many factors such as berry number, size, shape, and spatial location, which all relate to the rachis ramification patterns [3]. While certain features of cluster architecture can be discerned by looking at the cluster contour, a more precise analysis requires the identification and spatial localization of the individual berries within the cluster. Cluster architecture and compactness are determined genetically, as many genomic regions have been associated with trait variation [2–6]. However, environmental factors such as temperature, humidity, nutrient availability, and vineyard management, among others, are known to alter cluster architecture and compactness directly or indirectly [1,2,7,8].

Understanding the factors that influence cluster architecture and compactness, and to what extent they do so, has implications for vineyard management, breeding, and genetics research. For example, high cluster compactness has been associated with increased susceptibility to *Botrytis* bunch rot caused by *Botrytis cinerea* [9–11]. This, in turn, has implications in terms of vineyard management and cultivar preference, since fungicide applications can better reach berries within the cluster in the case of a more open, looser cluster. Furthermore, there is a greater temperature variability between the inner and outer berries in densely compacted clusters, impacting the maturation rate [8]. Additionally, restricted sun exposure to berries has been observed to intensify powdery mildew infections [12], thereby influencing fungicide application scheduling.

Exploring cluster architecture and compactness has been the focus of several studies utilizing qualitative and quantitative methods. Among qualitative approaches, researchers primarily rely on the OIV descriptors, a set of definitions established by the International Organization of Vine and Wine. For instance, the descriptor OIV 204, which addresses cluster density or compactness, categorizes grape clusters into five classifications ranging from very loose to very dense. Similarly, cluster architecture can be described using a combination of OIV 208 - bunch shape (cylindrical, conical, and funnel-shaped) and OIV 209 - number of wings of the primary bunch (ranging from 1 to 6 or more). Classifying clusters based on OIV descriptors often involves considering multiple characteristics simultaneously, which, while providing a comprehensive assessment, can be challenging to replicate and scale. For example, Richter et al. (2019) studied an F1 mapping population derived from crossing GF.GA-47-42 and Villard Blanc. Their study involved manually recording individual cluster and berry traits (e.g., berry number, cluster weight, rachis size and architecture, and shoulder length, among others), which accounted for approximately half of the observed variation compared to using the OIV 204 descriptor alone. This emphasizes the complexity of cluster architecture and how it is influenced by various individual characteristics, including cluster compactness.

Computer vision approaches can also be used to analyze cluster architecture and compactness. In this case, available methods involve 2D image analysis and 3D modeling, which all have the capability of producing quantitative traits. In many cases, the utilization of quantitative traits derived from these imaging approaches has proven to be more effective in genetics research and breeding compared to categorical traits [13]. Depending on the algorithm, some studies have focused on berry detection while others only on whole cluster analysis. For example, conventional segmentation on cluster images generated in the lab has been used to assess berry color and cluster architecture [4,14]. Cluster images generated directly in the vineyard have been used also for cluster identification and yield estimation using a variety of methods, however, prediction accuracy has varied because of challenging light conditions or occlusion [15–17]. Identifying and localizing berries within the cluster is crucial for determining cluster architecture and compactness. In this context, several approaches have been tested, including robotic laser scanning systems to reconstruct 3D representations of clusters and generate precise data regarding the 3D location of berries in a cluster [18]. Likewise, X-ray tomography has been employed to scan grapevine inflorescences, model berry growth, and infer phylogenetic relationships [19]. Partial 3D models of grape clusters have also been generated using stereo-vision, which in turn, allows berry counting [20]. Some other methodologies allow the estimation of berry numbers from images taken directly in the field. For example, in [21], the model developed allowed for an accurate prediction of berry counts in Niagara grapes, which are generally larger than most table and wine grapes. Neural networks have also been applied for berry segmentation and counting, and although they produced very accurate estimates, they were only used on very immature clusters with limited berry growth, low compactness, and sufficient contrast between berries [22].

Many of the image analysis-based methods used to describe cluster architecture and compactness relied on traditional segmentation methods. These methods often depend on labor-intensive, customized functions, manually engineered features, and error-prone thresholding designed for specific scenarios. As an alternative, deep learning models for image analysis, with their ability to capture latent image features, have shown promise across various fields, including medicine, surveillance and security, agriculture, biometrics, environmental sciences, and remote sensing, among others. However, these models are typically designed and trained for specific segmentation tasks, and unfortunately, their performance may significantly deteriorate when applied to new tasks, different image types, or varying external conditions. Large-scale foundational models have revolutionized artificial intelligence due to their remarkable zero-shot and few-shot generalization capabilities across a broad spectrum of downstream tasks [23,24]. Foundation models are neural networks trained on vast datasets using innovative learning methods and prompting objectives that generally do not require conventional supervised training labels, which makes them adaptable to a variety of external conditions [25]. Segment Anything Model (SAM) is a new foundation model that can be used as a zero-shot segmentation method [26]. SAM can be used out-of-the-box to segment a variety of objects in an image, or can be fine-tuned for a specific task, as the very recently developed MedSAM [27]. SAM was built on the largest segmentation dataset to date, with over 1 billion segmentation masks [26]. To segment an object, SAM requires the user to provide a prompt, which can take the form of a single point, a polygon (similar to a mask), a bounding box, or just text [25].

In this study, we demonstrated the capabilities of SAM to segment grape berries from 2D cluster images without additional model training or fine-tuning. Our research focused on four main aspects: 1) measuring the accuracy of SAM in identifying visible berries within a cluster image; 2) predicting hidden berries in a cluster image and assessing the impact of cluster imaging angle; 3) developing new quantitive methods to describe cluster architecture based on berry distributions within the clusters; and 4) assessing the repeatability of cluster architecture and compactness traits in replicated experiments.

**Materials and Methods**

**Plant material**

Cluster images obtained from an F1 mapping population (n=139 genotypes) derived from crossing Cabernet Sauvignon and Riesling were used to test SAM. Both Cabernet Sauvignon and Riesling, major wine grape cultivars around the world, display contrasting cluster architectures. Cabernet Sauvignon clusters are small to medium in size, conical, loose to well-filled, and with medium-long peduncles. Its berries are small, round, and blue-black. Riesling has smaller clusters, which can be cylindrical or globular, and sometimes winged; clusters are compact and with short peduncles. Riesling berries are small, round, and have a white-green skin coloration. This F1 progeny segregates for the traits mentioned above, making it an ideal candidate to evaluate the proposed pipeline. This population was planted in UC Davis Experimental Station in Oakville, Napa County, CA, USA (38°25′45.4″ N; 122°24′36.4″ W), in 2017. Vines were arranged using a randomized complete block design with three blocks and three vines per experimental unit. For this study, one vine per experimental unit was sampled (the one in the middle). For each vine, five representative clusters were imaged as described below.

**Image capture**

Five representative clusters per vine were imaged using the setup shown in Fig. S1. The setup included a reference circle to normalize measurements and account for potential variation in the location of the camera relative to the cluster. The camera used was a Canon EOS 70D with a 24mm prime lens, aperture of f/5, and exposure time of 1/500 sec. Images were 2472 x 3648px (~9 Mpx). All clusters were imaged from at least one angle. In addition, all the clusters from a subset of 99 vines were imaged from three additional angles (90, 180, and 270°). The latter was used to assess complex architectures that result from the presence of cluster ramifications or wings, and that are visible only from specific view angles.

A second image dataset was generated to validate the SAM algorithm. This dataset consisted of cluster images, each one accompanied by an image of all the individual berries detached and individually placed on a white surface (Fig. S2).

**Model and processing pipeline**

The images described above, without editing their original brightness or contrasts, were used as input for SAM. To reduce the amount of pixels to be processed, a region of interest (ROI) was manually defined, as indicated in Fig. 1. The pretrained ViT-H (Huge Version) image encoder was used for the segmentation phase (Checkpoint available at https://dl.fbaipublicfiles.com/segment_anything/sam_vit_h_4b8939.pth). The mask prediction was executed by applying the Automatic Mask Generator to the input, which was defined as the pixels within the ROI and a prompt, described as an XY grid of points equally distributed across the ROI. Different grid configurations, including 64 x 64, 128 x 128, and 256 x 256 were explored and tested for efficiency. After preliminary tests, the 128 x 128 grid captured all the grape masks without unnecessary computational overhead. As a preliminary analysis, SAM was executed using GPU, MPS, and CPU platforms to compare any potential segmentation differences, however, only computation time was affected. The output produced by SAM comprised bounding boxes in XYWH format, area, predicted intersection over union (IoU), stability scores, and mask segments formatted as COCO Run Length Encoding (RLE). The implementation of SAM, including ROI identification and automatic mask generation, was implemented in Python 3.11. The hardware tested was a g3.4xlarge AWS instance (single GPU, 16 GB RAM) and a System76 workstation (32 CPU, 256 GB RAM). Details on specific dependencies are available in the following GitHub repository: https://github.com/diazgarcialab/SAM-cluster-segmentation.

The RLE mask segments were decoded using pycocotools (https://github.com/cocodataset/cocoapi/blob/master/PythonAPI/pycocotools/mask.py) to derive the x and y coordinates of the mask contours and their position within the cluster. These coordinates were analyzed using the R package Momocs [28] to compute various parameters such as berry area, length, width, aspect ratio, perimeter, and color (represented as median red, green, and blue values). SAM is a segmentation tool rather than a classifier. As such, the segmented masks it produces may include, in addition to berries, other objects such as the clamp used to hold the clusters or the reference circle for size normalization. These objects can be easily identified and distinguished from berries due to their contrasting morphology and size, as described below. More often, some masks may encompass two or more berries, which were addressed using the IoU estimates. IoU is a metric used to evaluate the overlap between two bounding boxes or masks, commonly employed when assessing the accuracy of image segmentation models. In this study, IoU was calculated by determining the size of the overlapping region between two masks detected by SAM. For example, in instances where an overlapping mask covers two berries, each with its own mask, the overlapping mask will exhibit a larger size and IoU. Furthermore, filters based on criteria such as area, perimeter-to-area ratio, and aspect ratio were implemented to exclude objects other than berries. To refine the segmentation further, we employed a filtering approach using Elliptical Fourier Descriptors (EFD) and Principal Component Analysis (PCA) to eliminate non-berry objects, especially rachis parts. Initially, the x and y coordinates of objects were transformed into an 'Out' object using Momocs software, which facilitated the computation of EFD harmonic coefficients. These coefficients were then analyzed using PCA for visualization purposes, and outliers were identified through five rounds of outlier detection. Each round involved recalculating the harmonics and principal components with a cleaner dataset, adopting a threshold of ±2 standard deviations among the first ten principal components.

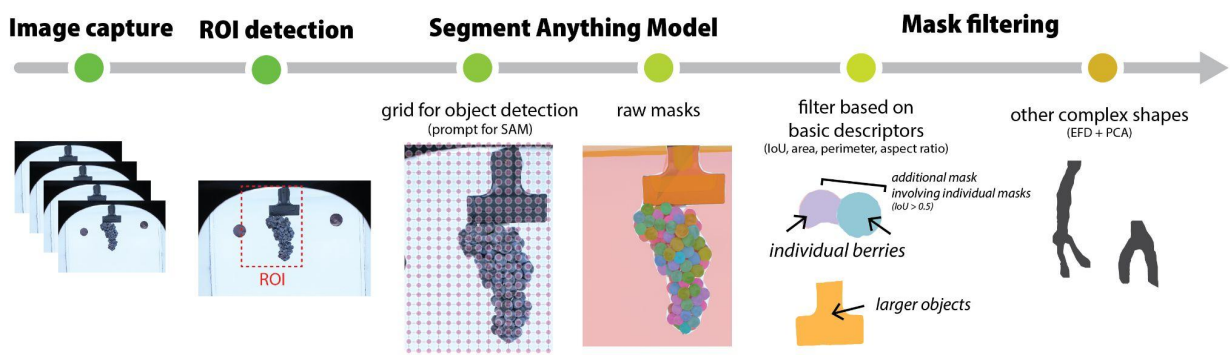

**Fig. 1 illustrates the summary of the pipeline employed for generating and processing SAM masks.** The process for each image is detailed below. Firstly, the region of interest (ROI) housing the cluster is identified. Subsequently, a grid of points separated by 128 x 128 pixels is utilized as input for object identification in SAM. Following this, masks undergo analysis based on various parameters including intersect over union (IoU), area, perimeter, length, width, aspect ratio, and Elliptical Fourier Descriptors (EFD) to discern non-berry objects.

Results

**Characteristics of the model implementation and implementation time**

The implementation of SAM on a population of 387 vines and 1,935 different clusters resulted in 215,090 masks. For 99 of the 387 vines, all the clusters were imaged four times, each time at a different angle (0, 90, 180, and 270°), which resulted in 3,431 cluster images. The identified masks included, among other things, individual berries, two or more berries, the clamp used to hold the clusters in place, stains/decolorations in the background, the reference circle for size normalization, and rachis segments. This outcome is expected as SAM utilizes an algorithm for unsupervised object segmentation, and not classification, within an area of interest defined by the user. As a result of the filtering, 32,425 masks containing two or more berries were removed using the intersection over union parameter, or IoU. Furthermore, since berries had an expected size and aspect ratio, 23,125 masks with significantly larger areas or aspect ratios, or located far from the cluster (stains in the background) were filtered out. Finally, the rest of the mask contours were analyzed with Momocs [28] using a combination of EFD and PCA, leading to the identification of 5,601 objects other than berries. After this filtering step, the number of true berry masks was 153,939 (61,151 masks discarded). Each cluster had, on average, 44.87 berries (median=42). Berry number varied between 5 and 130, and variation showed a normal distribution (Fig. S3).

Regarding computer time, about 720 cluster images per minute were processed by SAM using the computer setup described in the Materials and Methods section.

**2D cluster representations predict berry number and cluster size**

Berry counts from clusters imaged at four different angles were compared with the number of berries determined manually. The "manual" determination of berries was conducted using two methods. The first involved humans counting visible berries in a subset of 100 images, and then comparing these counts with SAM predictions. The second involved processing additional images of 17 clusters where all the berries were detached and placed individually on a surface. The analysis of these images is straightforward since there is no touching among berries, and there exists good contrast between the berry and surface colors (Fig. S2). In addition to being used to determine the true number of berries, these images also allowed the comparison of berry size, assuming that the masks generated from isolated, uncompressed berries imaged from the top approximate well to the real size of a berry.

As shown in Fig. 2A, the SAM algorithm does a very good job finding and segmenting all the berries in the cluster, independently of the angle it is being imaged. The berries identified were fully visible, represented as circles, or partially visible (Fig. 2B). The correlation between the berry number determined by humans and the SAM prediction was 0.96 (Fig. S4). There was also good agreement between SAM berry number predictions and the number of berries calculated from images with the individual berries. However, there was a clear underestimation, which varied depending on the imaging angle (Fig. 2C). Overall, the underestimation was approximately 50% of the real number but linear. In symmetric clusters (e.g., cylindrical with no ramifications or wings), images from all four angles yielded similar berry counts. Conversely, clusters with wings, as they were only visible from specific angles, increased the berry count prediction. While the berry count was underestimated, a linear regression model of the form $y \sim \beta_0 + \beta_1 x$ was sufficient to adjust the prediction considerably well (adjusted $R^2$=0.8723), as long as the cluster with the maximum number of berries (from the four images taken at different angles) was used in the model.

Berry size (measured as projected berry area) was more challenging to predict (Fig. 2D). Predictions were mostly overestimations and varied significantly depending on the imaging angle. Most berries were between 120 and 150 mm$^2$, with just a few having smaller sizes (<100 mm$^2$). Studying clusters with more variation in berry size might be required to better assess the correlation for this trait. Similar to berry counts, a linear model was fitted using all cluster views available for each cluster. Since it appeared to be linear, the fitted values were consistent with the real size estimations (adjusted $R^2$=0.8457).

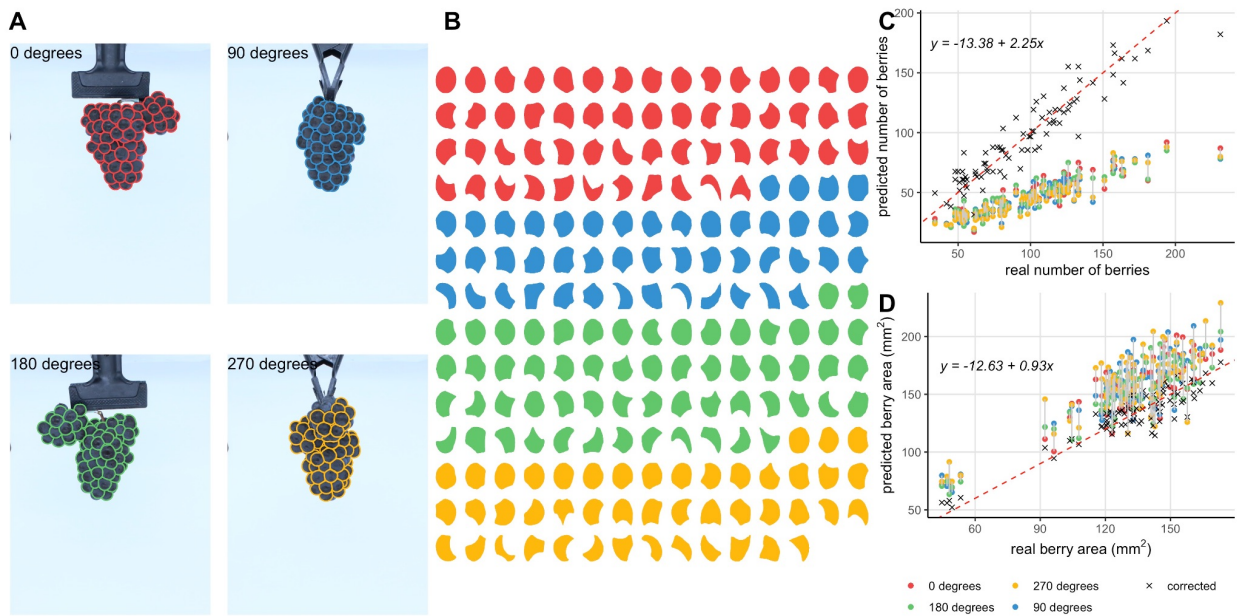

**Fig. 2. Prediction of berry number using SAM from cluster images. (A)** Identification of individual berries from four angles on the same cluster. **(B)** Berry masks

from cluster images in panel A, color-coded by angle view. **(C)** Correlation between real and predicted berry counts from SAM; predicted counts for each angle view in panel A are displayed. Points marked with an X represent corrected counts using the angle view with the maximum berries, adjusted with a linear model. **(D)** Correlation between real and predicted berry area; color and shape patterns are similar to panel C; corrected points were generated with a linear model of the form $y \sim \beta_0 + \beta_1 x$. The vertical red line indicates a one-to-one relationship between variables.

**Cluster angle matters**

Not all the berries in a cluster can be seen from a given angle; therefore, berry counts from 2D images were, as expected, underestimated (Fig. 2C). While cylindrical clusters are more common among cultivars, the presence of ramifications or wings, or other asymmetries, can impact the number of berries visible from a single view. To measure the effect of the image angle on the berry counts, 490 clusters from 99 vines were imaged from four different angles (0, 90, 180, and 270), and the berry counts and sizes were compared. In general, the berry count can vary by approximately ±50%, depending on the angle (Fig. 3A). As expected, opposing angles (0 and 180, 90 and 270) tend to have more similar results (Fig. 3B). In other words, when the cluster ramification or wing is fully visible from a given angle, it becomes invisible or hard to distinguish when the cluster is rotated 90°, and becomes fully visible again after another 90° rotation. Berry size was less dependent on the viewing angle (Fig. 3C). In general, berry size varied by +30%. The extent of the variation in berry count as a function of viewing angle is shown in Fig. 2D.

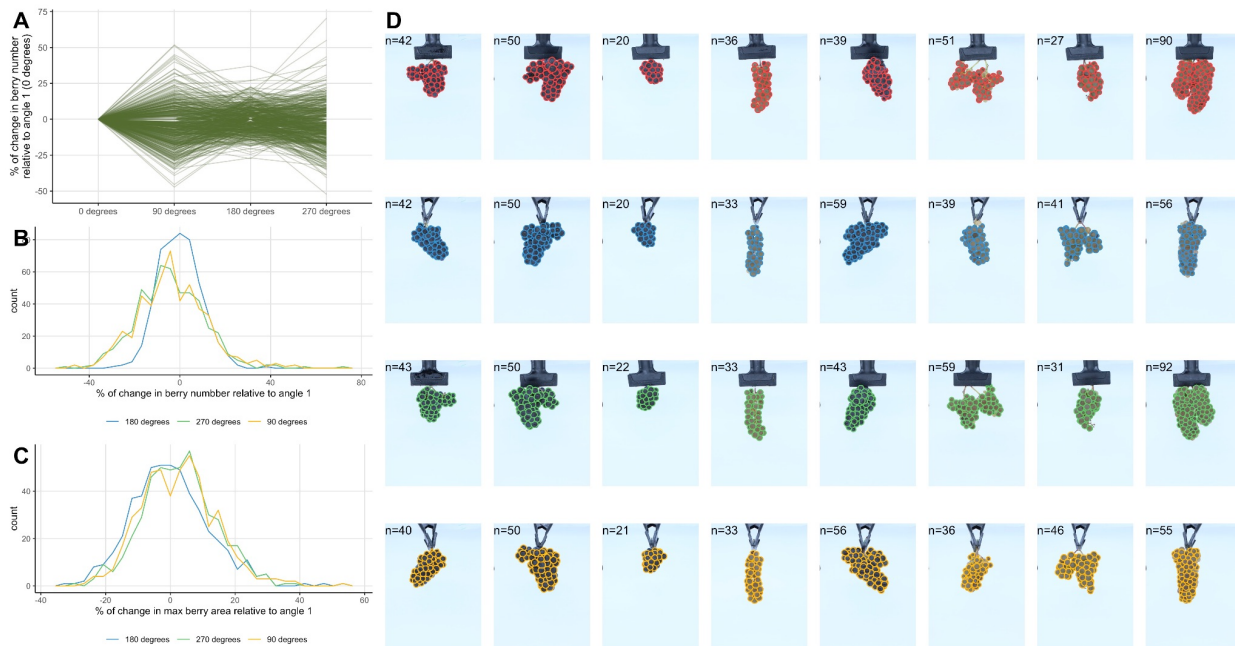

**Fig. 3. Impact of imaging angle on cluster analysis. (A)** Change in berry number relative to angle 1 (0°, first image); each green line represents a cluster imaged at four different angles. **(B)** Frequency plot of changes in berry number relative to angle 1, similar to panel A. **(C)** Frequency plot of changes in max berry area relative to angle 1. **(D)** Examples illustrating the effect of berry angle on SAM-detected berry counts; each column represents a different cluster, and each row represents a different angle (0, 90, 180, and 270°). The number of detected berries is indicated in each image. The first four clusters show little variation, while the last four exhibit extreme berry count variation.

**Cluster architecture**

A typical approach for measuring cluster architecture and compactness is based on whole cluster segmentation instead of berry segmentation (e.g., Underhill et al. 2020). While this method provides insightful information and is easy to implement, it ignores the spatial distribution of berries within the cluster. Moreover, in the setup used for photographing clusters, it is common to use clamps, hooks, or clips to hang clusters, which can then be challenging to identify during image analysis or post-processing. In those cases, a common strategy is to crop the top of the image to remove such objects. When the peduncle is long, cropping the image does not affect the analysis; however, in clusters with short peduncles or prominent shoulders, cropping the image results in cropped berries as well. In this study, although the clamps (and other objects) were masked by SAM, because of different colors, sizes, and shapes, they were easy to identify and remove.

To illustrate the capabilities of cluster architecture analysis using berry locations, empirical cumulative distribution functions were developed along the y (from the top of the cluster, or the peduncle, to the bottom, or cluster tip) and x axes (from left to right). The distribution functions provided different levels of information. For example, they allowed the estimation of symmetry along both the x and y axes. With these symmetry estimators, cylindrical or globular clusters are expected to have a more uniform cumulative distribution. On the other hand, clusters with ramifications or significant ramifications will show a cumulative distribution along the x-axis skewed opposite to the main ramification.

A cluster with a prominent wing, photographed from different angles, is provided as an example in Fig. 4A. At 0 and 180° views, the wing is not visible, as it is either in front and completely aligned with the main cluster, or in the back. In this case, the cluster appears more cylindrical and symmetrical along both axes. The empirical cumulative distribution functions for these two views, shown as red and green dots in Fig. 4B and 4E, were more uniform and appeared as straight diagonal lines. Conversely, at 90 and 270° views, the wing becomes visible and produces a very skewed distribution along the x-axis. Since the 90 and 270° views, and the 0 and 180° views, can be seen as "mirror" images, the distribution functions in Fig. 4D and 4E also display this mirroring feature.

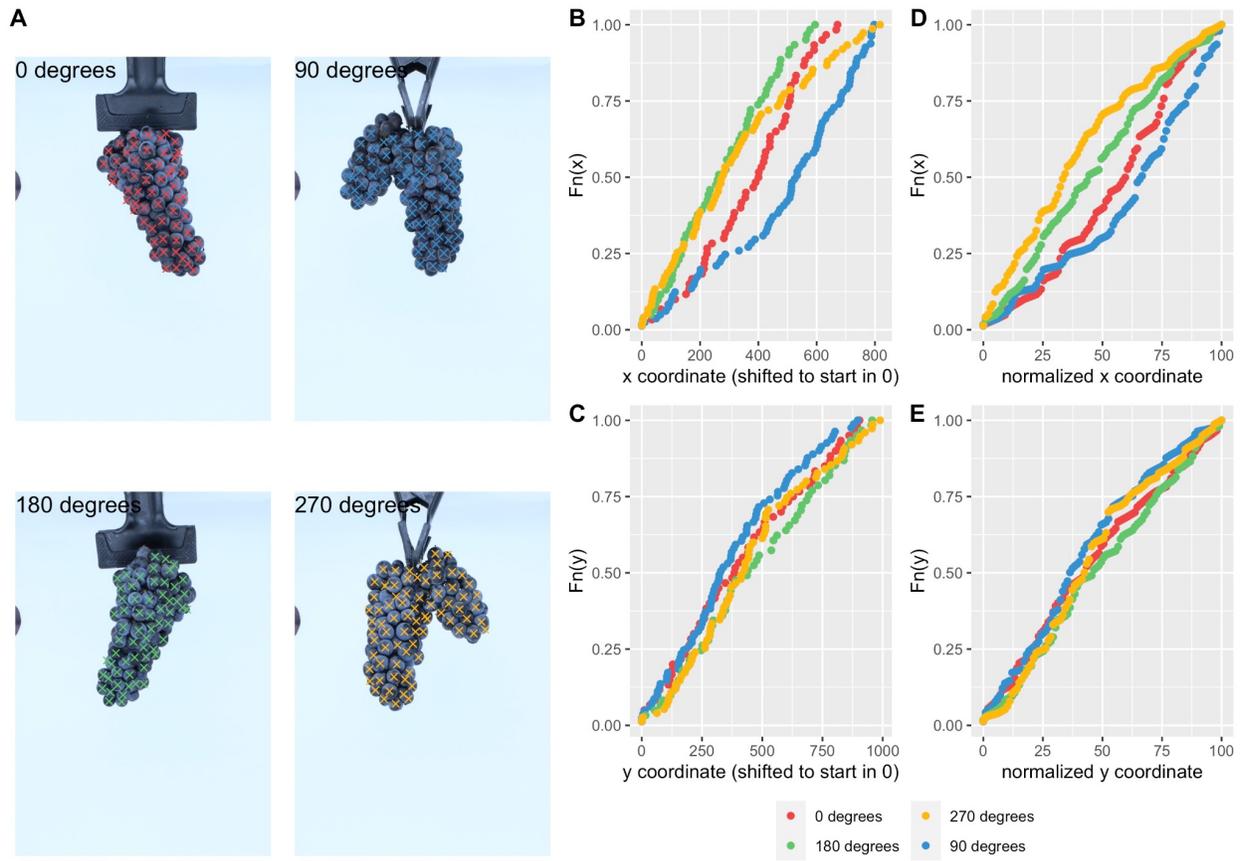

**Fig. 4. Cumulative distributions of berry locations along the horizontal and vertical axes. (A)** Example of berries identified in a cluster imaged from four different angles (0, 90, 180, and 270°). **(B)** Empirical cumulative distributions along the x-axis for the four angle views; berry locations along the x-axis are shifted to start at 0. **(C)** Similar to panel B, but for the y-axis. **(D)** Similar to panel B, but berry locations along the x-axis are scaled from 0 to 100 and sampled with n=100. **(E)** Similar to panel D, but for berry locations along the y-axis.

Masks generated by SAM for each berry object were represented as x, y coordinates, and their corresponding polygons were drawn, as shown in Fig. 2A and Fig. 3D. Combining all the berry polygons produced a representation of entire clusters. When a cluster has a cylindrical or globular shape, and no wings are present, representing its shape is simple. However, when other cluster features are present, such as wings, shoulders, and conical forms, among others, the so-called cluster shape descriptor can vary depending on how detailed these complex features are represented.

For example, for a cluster with a prominent wing, as the one shown in Fig. 4A, should the outline (or contour) defining the cluster shape include the sinus formed by the two wings? If so, how far inside the sinus? The opposite approach would be to simply connect the tips of the wing and the main cluster formation, which would produce a simpler polygon. The same applies to the presence of shoulders and curvatures along the cluster. Fig. 5 illustrates the same eight clusters from Fig. 3D, outlined using concave hulls with varying degrees of detail, from top to bottom. At the top panels, the cluster outlines preserved detailed features such as shoulders, indentations, separations between wings, etc. Towards the bottom part of the figure, most of these features were lost.

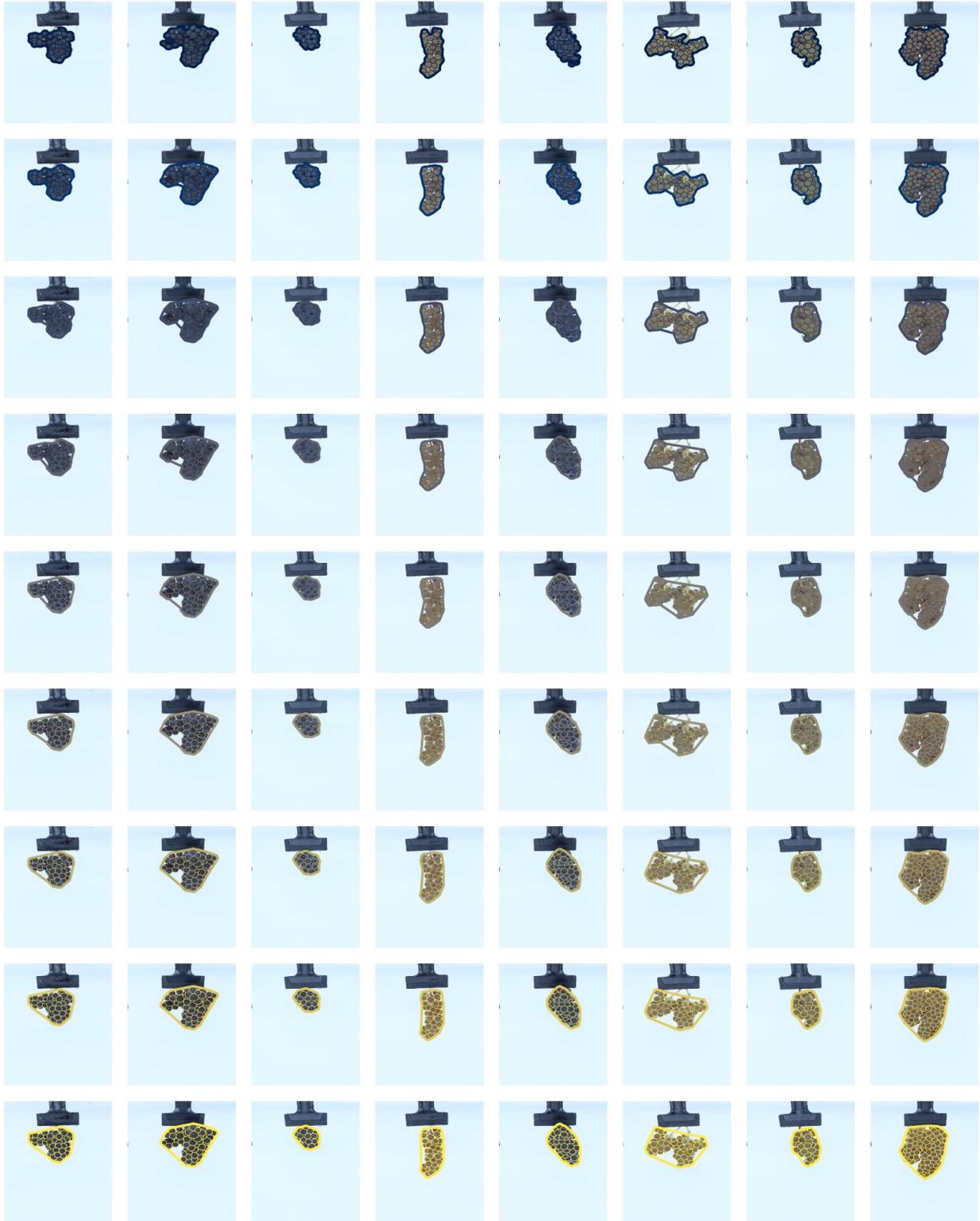

**Fig. 5. What is cluster architecture?** Example of concave hull calculation for different clusters (in columns) at different cluster shape definition levels (from top to bottom,

higher to lower definition); concave hulls are calculated on the union of all berry masks in the cluster.

The approaches described above to measure cluster architecture were applied to all 3,431 cluster representations analyzed in this study. Cumulative distribution functions for axes x and y showed varying levels of asymmetry. Along the x-axis (Fig. 6A), the asymmetry is due to having more berries either on the right or the left side of the cluster (likely because of the presence of a wing). For example, the green lines in Fig. 6A represent the distribution functions of clusters in which more berries exist on the left side of the cluster (as much as ~75%). Conversely, the purple lines represent clusters with a larger accumulation of berries on the right side of the cluster. Finally, gray lines represent more symmetrical clusters, with an equal amount of berries on the left and right. Regarding the y-axis (Fig. 6B), most of the asymmetry is towards the base of the cluster, which is expected, as many clusters exhibit conical forms. Importantly, the color assignments (i.e., categories) in Fig. 6A and 6B are subjective and for illustrative purposes only.

Then, cluster shape variation was studied using the polygons generated using concave hulls. The concave hulls were generated using a conservative level of cluster feature preservation (using the function R function *sf::st_concave_hull()*, with *ratio=5*) but with enough resolution to capture major asymmetries, wings, and shoulders. In general, cluster shape exhibited a continuous gradient of variability with no clear group formation (Fig. 6C and 6D). In other words, there were no groups formed only with, for example, winged and non-winged clusters, or symmetric and non-symmetric clusters. Instead, asymmetries can be either small and slightly visible, and increase gradually in size and separation from the main cluster. To understand what cluster features were associated with each PC, 100 clusters with extreme PC scores (50 more negative and 50 more positive) were plotted for PCs 1-4 (Fig. 6E). PC1, which explained 53.23% of the variation, was associated with aspect ratio, with more circular/globular clusters having more negative values, and very elongated clusters with more positive values. PC2, which explained 18.29% of the variation, was associated with the location of the asymmetries along the x-axis (either to the left or the right). Finally, both PCs 3 and 4, which accounted for a little less than 18%, explained other more complex features (wings and shoulders) that are more difficult to discern.

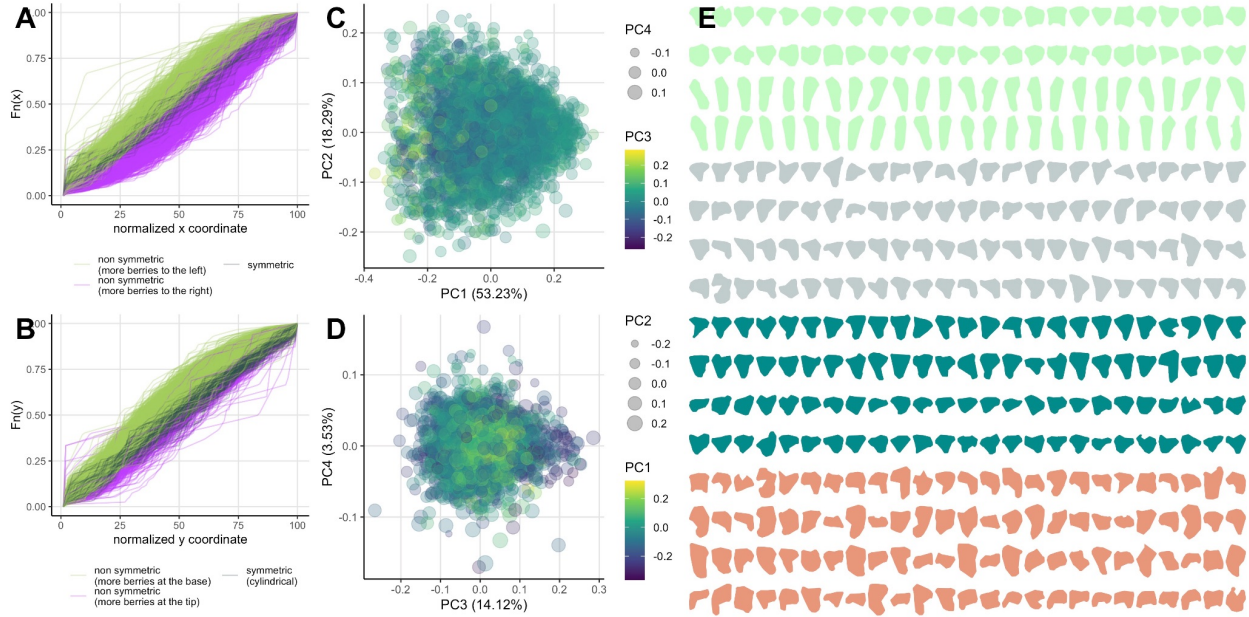

**Fig. 6. Comprehensive analysis of cluster architecture using cumulative distribution function and PCA of concave hulls.** Empirical cumulative distributions for 3431 clusters using berry locations along the **(A)** x and **(B)** y axes; berry locations along both x and y axes are scaled from 0 to 100 and sampled with n=100, similar to Fig. 3D and E. In both cases, the green lines correspond to distributions with a normalized coordinate 25 larger than 0.3 and a normalized coordinate 75 larger than 0.8; the purple lines have a normalized coordinate 25 < 0.2 and a normalized coordinate 75 < 0.7; finally, the gray lines have a coordinate 25 between 0.2 and 0.3, coordinate 50 between 0.45 and 0.55, and coordinate 75 between 0.7 and 0.8. Variation in cluster architecture along principal components 1 and 2 **(B)** and 3 and 4 **(C)**. In panel C, different colors and sizes correspond to variations in principal components 3 and 4, respectively. Similarly, in panel D, point color and size correspond to variations in principal components 1 and 2, respectively. **(E)** One hundred clusters sampled from the extremes of principal components 1 (green), 2 (gray), 3 (dark cyan), and 4 (salmon); the clusters in each color group are ordered from left to right and by rows according to their corresponding principal component values.

**Is the level of sensitivity to complex cluster features meaningful?**

The methodologies employed in this study for identifying berries within a cluster, counting them, studying their spatial distribution to generate cumulative distribution functions, and applying PCA to examine cluster shape variation demonstrated high sensitivity (Fig. 6). However, a critical question is: are these features primarily driven by genetic variation, or are they simply a result of environmental and non-genetic factors?

The primary aim of this research was to implement SAM for berry identification and propose methodologies for leveraging this information in cluster architecture and compactness analysis. Therefore, the focus was not on characterizing specific cultivars or genotypes in the surveyed population but rather on sampling diverse cluster variations. Nevertheless, as mentioned earlier, the sampled vines are part of a mapping population between Riesling and Cabernet Sauvignon, planted in a randomized complete block design with three contiguous vines per genotype per block. This design allowed the calculation of repeatability, expressed as the percentage of genetic variance relative to the phenotypic variance.

First, to assess the consistency of the phenotypes measured in this study, boxplot graphs per genotype were examined for 18 variables. These variables included basic descriptors such as berry count, area, length, and width, all computed from the berry masks identified by SAM. Additionally, cluster compactness was calculated as the ratio between the sum of all berry areas and the concave hull area. Using the empirical cumulative distribution functions, the predicted percentage of berries at x or y=25, 50, and 75 was also determined. In terms of cluster architecture based on concave hulls, PCs 1 and 2 were included. Finally, cluster length, width, perimeter, and aspect ratio were computed using the concave hulls.

Overall, variables such as berry count, area, length, and width, as well as cluster area, length, width, and perimeter, showed good consistency (Fig. 7A), high correlation (Fig. 7B), and medium-to-high repeatability (Fig. 7C). While descriptors derived from cumulative distributions showed a correlation among themselves, except for ECDF at x=25 and y=25, their variability was higher, likely influenced by non-genetic sources given their very low or zero repeatability. Cluster compactness demonstrated little correlation with other traits but exhibited good consistency with a repeatability of ~0.6. PC1 from the PCA conducted on concave hulls, and related to cluster aspect ratio, also showed good consistency and medium to high repeatability. In summary, these analyses revealed that many variables computed from the berry masks identified by SAM, along with others describing more complex features in the cluster, possess a genetic component. Nevertheless, certain variables, particularly those originating from empirical cumulative distribution functions, seem to be strongly affected by variations in the environment.

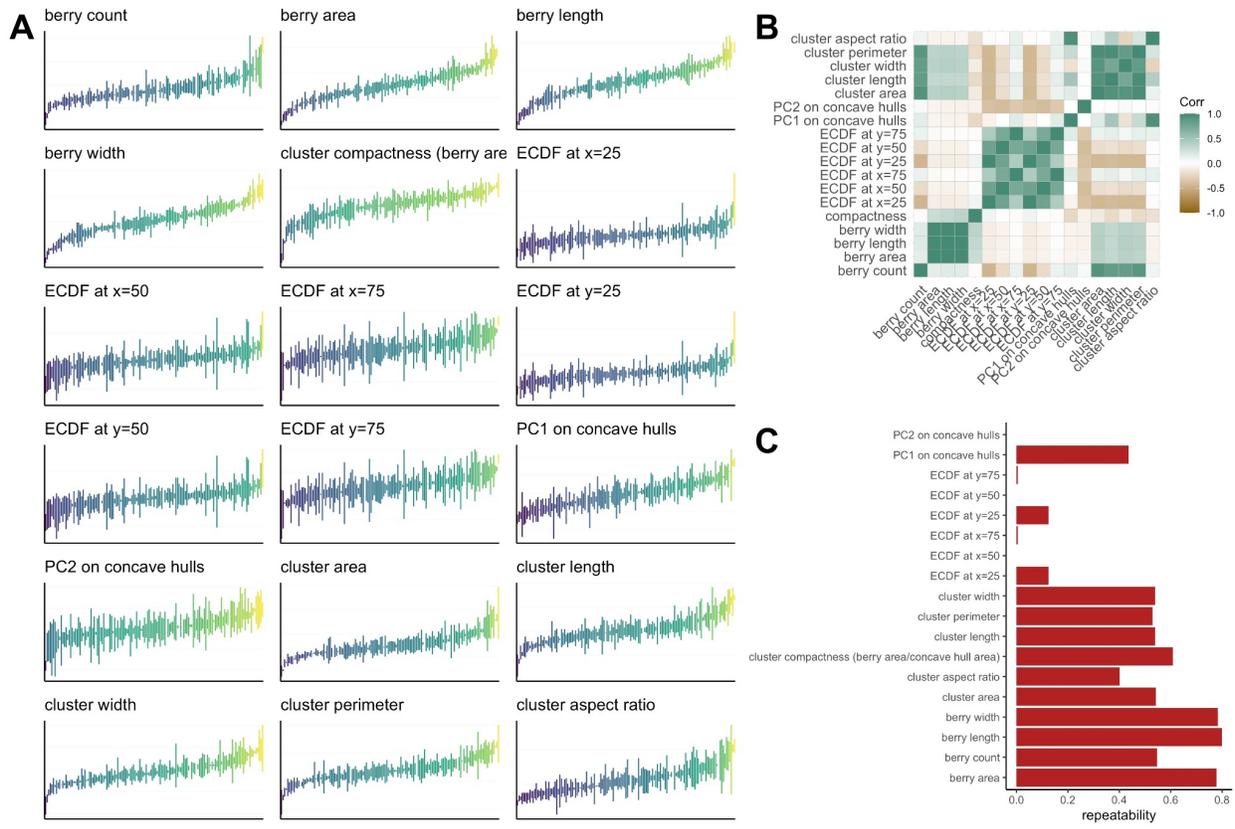

**Fig. 7. Variability in berry and cluster characteristics. (A)** Variation in berry characteristics and cluster architecture grouped by genotype; each genotype was replicated in three blocks; for each replicated vine, five clusters were sampled and imaged from one or four angles; genotypes are ordered by mean value, and names are omitted due to space constraints. **(B)** Pearson's correlation between traits. **(C)** Repeatability is calculated as the proportion of genetic variance relative to the phenotypic variance.

## Discussion

Several computational, image-based strategies have been implemented to measure grapevine cluster architecture and compactness. However, only a few have been utilized for identifying individual berries within clusters (Schöler and Steinhage 2015; Li et al. 2019; Ivorra et al. 2015; Luo et al. 2021). Most of these strategies rely on non-generalizable mathematical and analytical frameworks for analyzing colored images. Humans can easily discern individual berries in a cluster image, even when taken in the field or under challenging light conditions. Therefore, it is reasonable to assume that machine learning algorithms could achieve similar capabilities. However, until now, these models have primarily been applied to cluster identification rather than berry identification. This does not discard the potential use of "conventional" deep learning approaches trained with human-segmented berries, but they would require a substantial amount of image labeling for training.

With the recent introduction of foundation models, particularly the Segment Anything Model, or SAM (Kirillov et al. 2023), objects of interest can be automatically segmented without the need for additional training or fine-tuning, at least for natural objects. In some specific cases, such as in medical imaging, additional fine-tuning allows for more accurate predictive models capable of analyzing many different image types [27]. Here, we demonstrate that out-of-the-box SAM can accurately segment berries in a 2D grape cluster image with up to a 0.96 correlation (human berry counts vs. SAM predictions on visible berries in 2D images; Fig. S4). While one might argue that the segmented masks produced by SAM in this study needed supervised classification to identify berry objects exclusively, the implementation of filters (IoU, size, area, EFD, and PCA) was straightforward. This approach can be applied to hundreds, thousands, or even millions of masks without any changes to the programming. A continuation of this work could be the development of an automatic classifier based on, for example, YOLO, that can use cropped images based on bounding boxes generated by SAM.

In terms of computation time, each image was processed in about 30 seconds, which is sufficient for certain applications, like cultivar description or genetics research (e.g., genetic mapping). For large-scale applications such as real-time cluster analysis under field conditions, other SAM-like implementations such as EdgeSAM [29], fastSAM [30], and EfficientSAM [31], could be adopted.

Although the number of berries visible in a cluster image underestimates the actual number of berries in a cluster, this underestimation can be corrected using a linear regression model (Fig. 2C). Moreover, to compensate for the variability in berry number caused by cluster ramifications of rachis visible only from certain angles, an additional image, for example, taken at 90°, can correct for any berry count underestimation. Notably, the berry masks generated by SAM can be used for comprehensive cluster architecture analysis, which is only possible if berries are spatially located.

While some of the trait variation in cluster architecture and compactness, particularly that captured by the analysis of empirical cumulative distributions, was influenced by environmental factors, these traits can still have applications in determining vineyard management practices. For instance, cluster thinning and/or tipping could be targeted towards asymmetric or winged clusters, or those with specific architectures. In table grape production, certain cluster architectures might be more appealing to consumers [32]. For these types of applications to be feasible under field conditions, SAM would have to be integrated into an existing pipeline that processes field images obtained with cameras mounted on rovers or tractors. In [16], for example, images from vines were acquired using a sensing kit equipped with RGB cameras, and further processed using YOLO to identify clusters within the image for yield prediction. In this case, SAM could be incorporated into this pipeline to compute additional variables regarding berry count, size, and cluster architecture.

The observation that the cumulative distribution functions (Fig. 6A and 6B) explaining cluster architecture showed lower or zero repeatability is specific to the Riesling by Cabernet Sauvignon population analyzed in this study. However, this does not rule out the possibility that other mapping or breeding populations display heritable variation for these traits. Consequently, these traits could still be valuable for genetics research or selection purposes in other mapping or breeding populations.

## Acknowledgments

The authors would like to thank Guillermo Garcia Zamora, Veronica Nunez, Jose Munoz, Sadikshya Sharma, Yaniv Lupo, Hollywood Banayad, and Dan Ng for their support during vineyard management, harvest, and image annotation

This project was partially supported by USDA-NIFA Specialty Crop Research Initiative Award No. 2022-51181-38240.

## Author contributions

ETL developed the proof of concept and set up the computational workflow to implement SAM. ETL and LDG conceived and designed the field experiment. JLV and GGZ supported fieldwork and cluster imaging. ETL and LDG and wrote the manuscript.

## Competing interests

The authors declare that there is no conflict of interest regarding the publication of this article.

## Supplemental Materials

All supplementary figures are available in the supporting file.

## References

1.  Tello J, Ibáñez J. What do we know about grapevine bunch compactness? A state-of-the-art review. Aust J Grape Wine Res. 2018;24: 6–23. doi:10.1111/ajgw.12310

2.  Richter R, Gabriel D, Rist F, Töpfer R, Zyprian E. Identification of co-located QTLs and genomic regions affecting grapevine cluster architecture. Theor Appl Genet. 2019;132: 1159–1177. doi:10.1007/s00122-018-3269-1

3. Correa J, Mamani M, Muñoz-Espinoza C, Laborie D, Muñoz C, Pinto M, et al. Heritability and identification of QTLs and underlying candidate genes associated with the architecture of the grapevine cluster (Vitis vinifera L.). Theor Appl Genet. 2014;127: 1143–1162. doi:10.1007/s00122-014-2286-y

4. Underhill A, Hirsch C, Clark M. Image-based Phenotyping Identifies Quantitative Trait Loci for Cluster Compactness in Grape. J Am Soc Hortic Sci. 2020;145: 363–373. doi:10.21273/JASHS04932-20

5. Fanizza G, Lamaj F, Costantini L, Chaabane R, Grando MS. QTL analysis for fruit yield components in table grapes (Vitis vinifera). Theor Appl Genet. 2005;111: 658–664. doi:10.1007/s00122-005-2016-6

6. Richter R, Rossmann S, Töpfer R, Theres K, Zyprian E. Genetic analysis of loose cluster architecture in grapevine. BIO Web of Conferences. 2017;9: 01016. doi:10.1051/bioconf/20170901016

7. Li-Mallet A, Rabot A, Geny L. Factors controlling inflorescence primordia formation of grapevine: their role in latent bud fruitfulness? A review. Botany. 2016;94: 147–163. doi:10.1139/cjb-2015-0108

8. Pieri P, Zott K, Gomès E, Hilbert G. Nested effects of berry half, berry and bunch microclimate on biochemical composition in grape. OENO One. 2016;50: 23. doi:10.20870/oeno-one.2016.50.3.52

9. Hed B, Ngugi HK, Travis JW. Relationship Between Cluster Compactness and Bunch Rot in Vignoles Grapes. Plant Dis. 2009;93: 1195–1201. doi:10.1094/PDIS-93-11-1195

10. Vail ME, Wolpert JA, Gubler WD, Rademacher MR. Effect of Cluster Tightness on Botrytis Bunch Rot in Six Chardonnay Clones. Plant Dis. 1998;82: 107–109. doi:10.1094/PDIS.1998.82.1.107

11. Vali ME, Marois JJ. Grape cluster architecture and the susceptibility of berries to Botrytis cinerea. Phytopathology. 1991;81: 188–191.

12. Austin CN, Wilcox WF. Effects of sunlight exposure on grapevine powdery mildew development. Phytopathology. 2012;102: 857–866. doi:10.1094/PHYTO-07-11-0205

13. Azevedo CF, Ferrão LFV, Benevenuto J, de Resende MDV, Nascimento M, Nascimento ACC, et al. Using visual scores for genomic prediction of complex traits in breeding programs. Theor Appl Genet. 2023;137: 9. doi:10.1007/s00122-023-04512-w

14. Underhill A, Hirsch CD, Clark MD. Evaluating and Mapping Grape Color Using Image-Based Phenotyping. Plant Phenomics. 2020;2020: 8086309. doi:10.34133/2020/8086309

15. Font D, Tresanchez M, Martínez D, Moreno J, Clotet E, Palacín J. Vineyard yield estimation based on the analysis of high resolution images obtained with artificial illumination at night. Sensors . 2015;15: 8284–8301. doi:10.3390/s150408284

16. Olenskyj AG, Sams BS, Fei Z, Singh V, Raja PV, Bornhorst GM, et al. End-to-end deep learning for directly estimating grape yield from ground-based imagery. Comput Electron Agric. 2022;198: 107081. doi:10.1016/j.compag.2022.107081

17. Nuske S, Wilshusen K, Achar S, Yoder L, Narasimhan S, Singh S. Automated visual yield estimation in vineyards. J Field Robot. 2014;31: 837–860. doi:10.1002/rob.21541

18. Schöler F, Steinhage V. Automated 3D reconstruction of grape cluster architecture from sensor data for efficient phenotyping. Comput Electron Agric. 2015;114: 163–177. doi:10.1016/j.compag.2015.04.001

19. Li M, Klein LL, Duncan KE, Jiang N, Chitwood DH, Londo JP, et al. Characterizing 3D inflorescence architecture in grapevine using X-ray imaging and advanced morphometrics: implications for understanding cluster density. J Exp Bot. 2019;70: 6261–6276. doi:10.1093/jxb/erz394

20. Ivorra E, Sánchez AJ, Camarasa JG, Diago MP, Tardaguila J. Assessment of grape cluster yield components based on 3D descriptors using stereo vision. Food Control. 2015;50: 273–282. doi:10.1016/j.foodcont.2014.09.004

21. Luo L, Liu W, Lu Q, Wang J, Wen W, Yan D, et al. Grape Berry Detection and Size Measurement Based on Edge Image Processing and Geometric Morphology. Machines. 2021;9: 233. doi:10.3390/machines9100233

22. Aquino A, Diago MP, Millán B, Tardáguila J. A new methodology for estimating the grapevine-berry number per cluster using image analysis. Biosystems Eng. 2017;156: 80–95. doi:10.1016/j.biosystemseng.2016.12.011

23. Zhang Y, Jiao R. Towards Segment Anything Model (SAM) for Medical Image Segmentation: A Survey. arXiv [eess.IV]. 2023. Available: http://arxiv.org/abs/2305.03678

24. Zhou C, Li Q, Li C, Yu J, Liu Y, Wang G, et al. A Comprehensive Survey on Pretrained Foundation Models: A History from BERT to ChatGPT. arXiv [cs.AI]. 2023. Available: http://arxiv.org/abs/2302.09419

25. Mazurowski MA, Dong H, Gu H, Yang J, Konz N, Zhang Y. Segment anything model for medical image analysis: An experimental study. Med Image Anal. 2023;89: 102918. doi:10.1016/j.media.2023.102918

26. Kirillov A, Mintun E, Ravi N, Mao H, Rolland C, Gustafson L, et al. Segment Anything. arXiv [cs.CV]. 2023. Available: http://arxiv.org/abs/2304.02643

27. Ma J, He Y, Li F, Han L, You C, Wang B. Segment anything in medical images. Nat Commun. 2024;15: 654. doi:10.1038/s41467-024-44824-z

28. Bonhomme V, Picq S, Gaucherel C, Claude J. Momocs: outline analysis using R. J Stat Softw. 2014;56: 1–24.

29. Zhou C, Li X, Loy CC, Dai B. EdgeSAM: Prompt-In-the-Loop Distillation for On-Device Deployment of SAM. arXiv [cs.CV]. 2023. Available: http://arxiv.org/abs/2312.06660

30. Zhao X, Ding W, An Y, Du Y, Yu T, Li M, et al. Fast Segment Anything. arXiv [cs.CV]. 2023. Available: http://arxiv.org/abs/2306.12156

31. Xiong Y, Varadarajan B, Wu L, Xiang X, Xiao F, Zhu C, et al. EfficientSAM: Leveraged Masked Image Pretraining for Efficient Segment Anything. arXiv [cs.CV]. 2023. Available: http://arxiv.org/abs/2312.00863

32. Zhou J, Cao L, Chen S, Perl A, Ma H. Consumer-assisted selection: the preference for new tablegrape cultivars in China. Aust J Grape Wine Res. 2015;21: 351–360. doi:10.1111/ajgw.12156